\documentclass[runningheads]{llncs}
\usepackage[T1]{fontenc}
\usepackage{graphicx,verbatim}

\usepackage{url}
\usepackage{esvect}
\usepackage[usenames,dvipsnames,table]{xcolor}
\usepackage{amsmath}
\usepackage{amssymb}
\usepackage{mathtools}
\usepackage[colorlinks=true, allcolors=blue]{hyperref}
\usepackage{algpseudocode}
\usepackage{algorithm}
\usepackage{setspace}
\usepackage{booktabs}
\usepackage{subcaption}
\usepackage{tabularray}
\usepackage{hyperref}
\usepackage{tikz}
\usepackage{multirow}
\usepackage{booktabs}
\usepackage[capitalise]{cleveref}
\usetikzlibrary{bayesnet, calc, arrows.meta, bending}
\usetikzlibrary{decorations.pathreplacing,angles,quotes,shapes.multipart, calligraphy}

\newcommand{\unintervened}[1]{\textcolor{gray}{#1}}

\usepackage{multirow}

\begin{document}
\title{Segmentor-guided Counterfactual Fine-Tuning for Locally Coherent and Targeted Image Synthesis}

\titlerunning{Segmentor-Guided Counterfactual Fine-Tuning for Image Synthesis}
%
%

\author{
Tian Xia\inst{1} \and
Matthew Sinclair\inst{1,2} \and
Andreas Schuh\inst{1,2} \and
Fabio De Sousa Ribeiro\inst{1} \and
Raghav Mehta\inst{1} \and
Rajat Rasal\inst{1} \and
Esther Puyol-Ant\'{o}n\inst{1,2} \and
Samuel Gerber\inst{2} \and
Kersten Petersen\inst{2} \and
Michiel Schaap\inst{1,2} \and
Ben Glocker\inst{1}
}
\authorrunning{T. Xia et al.}

\institute{
Department of Computing, Imperial College London, UK \\
\email{t.xia@imperial.ac.uk} \and
Heartflow, Inc., Mountain View, CA, USA 
}

\maketitle             
\begin{abstract}
Counterfactual image generation is a powerful tool for augmenting training data, de-biasing datasets, and modeling disease. Current approaches rely on external classifiers or regressors to increase the effectiveness of subject-level interventions (e.g., changing the patient's age). For structure-specific interventions (e.g., changing the area of the left lung in a chest radiograph), we show that this is insufficient, and can result in undesirable global effects across the image domain. Previous work used pixel-level label maps as guidance, requiring a user to provide hypothetical segmentations which are tedious and difficult to obtain. We propose \emph{Segmentor-guided Counterfactual Fine-Tuning} (Seg-CFT), which preserves the simplicity of intervening on scalar-valued, structure-specific variables while producing locally coherent and effective counterfactuals. We demonstrate the capability of generating realistic chest radiographs, and we show promising results for modeling coronary artery disease. Code: \url{https://github.com/biomedia-mira/seg-cft}.
\end{abstract}

\section{Introduction}
Causal questions, such as \textit{``How would this patient's disease have progressed if treatment A had been administered instead of treatment B?''}, are fundamental to scientific inquiry and clinical decision-making. Addressing such questions requires a causal framework capable of simulating realistic scenarios from observed data—going beyond conventional statistical models that primarily capture correlations. Causal models explicitly represent how changes in one factor influence another, enabling the exploration of both real-world interventions and hypothetical, or counterfactual, scenarios. The ability to generate counterfactual images has become particularly valuable in medical imaging, where it facilitates a range of applications, including data augmentation \cite{ilse2021selecting,roschewitz2024robust}, bias mitigation \cite{kumar2023debiasing}, explainability~\cite{pegios2024diffusion}, and disease progression modeling~\cite{puglisi2024enhancing}. By enabling targeted modifications to patient images, counterfactual generation can support model generalization, improve interpretability, and allow researchers to explore alternative clinical pathways. 

Recent efforts~\cite{louizos2017causal,kocaoglu2017causalgan,tran2017implicit,yang2021causalvae,sanchez2022diffusion,geffner2022deep}  have tried to integrate causality with deep generative models, including GANs~\cite{goodfellow2020generative}, VAEs~\cite{kingma2013auto}, and diffusion models~\cite{sohl2015deep,ho2020denoising,song2021scorebased}. However, most methods focus on association or intervention, without a principled approach to counterfactual reasoning—the highest level in Pearl’s causal hierarchy. Notable exceptions include Neural Causal Models (NCMs) \cite{xia2021causal,xia2023neural,pan2024counterfactual} and Deep Structural Causal Models (DSCMs) \cite{pawlowski2020deep,monteiro2023measuring,de2023high}, which integrate causal structures with deep generative models.

Ribeiro et al.~\cite{de2023high} proposed to train the generative causal model using a hierarchical variational auto-encoder (HVAE) conditioned on the assumed causal parents. However, relying solely on standard likelihood-based training was found to result in suboptimal axiomatic \textit{effectiveness}~\cite{monteiro2023measuring}, meaning that the generative model may fail to enforce counterfactual consistency—ignoring conditioning on intervened-upon parents in the forward model post-abduction~\cite{xia2024mitigating}. To address this, counterfactual fine-tuning (CFT) was proposed as an additional step, refining the HVAE with pretrained parent classifiers or regressors to improve adherence to causal structure~\cite{de2023high}. To this end, previous works~\cite{de2023high,xia2024mitigating,roschewitz2024robust,roschewitz2024counterfactual,ibrahim2024semi} focused on patient-level characteristics (e.g., sex, age, disease status). The effectiveness of DSCM and CFT has not been validated on structure-specific interventions, such as modifying specific anatomical regions or localized diease patterns. 

In this paper, we focus on these structure-specific interventions. We find that the previous CFT with regressors (Reg-CFT) is not sufficient for locally coherent and targeted counterfactual generation. To enable localized control of image generation, one potential approach is using segmentation masks to guide generative models~\cite{perez2024radedit,alaya2024mededit}. But integrating masks into a causal framework poses challenges: (i) defining their causal role remains unclear, and (ii) requiring pre-defined counterfactual masks reduces usability as these are difficult to obtain.

{Recent work has shown that medical image classifiers often rely on spurious or non-local features, motivating the use of spatial supervision via segmentation to improve specificity and robustness~\cite{hooper2023case,luo2022rethinking,saab2022reducing,aslani2022optimising}.}
Building on this insight, we propose Segmentor-guided Counterfactual Fine-Tuning (Seg-CFT), a method for fine-grained anatomical control in counterfactual image generation. We use scalar-valued variables (e.g., the area of the left lung) as guiding signals for counterfactual generation, which preserves the simplicity of the user interaction, avoiding any inputs at the pixel-level. In Seg-CFT, we utilise pre-trained, weight-frozen segmentors to increase the counterfactual effectiveness of DSCMs. The values for the intervened variables are directly determined from the obtained segmentations and compared against the desired user-specified target value in the loss function when fine-tuning the DSCM output. This enables a simple mechanism for intervening directly on structure-specific properties and, as we will show, yields locally coherent and targeted modifications of anatomical structures.

In summary, our key contributions are the following: 
\begin{itemize}
    \item We propose a novel guidance mechanism for counterfactual fine-tuning, which enables fine-grained structural control of localised interventions.
    \item We provide a comparative analysis against state-of-the-art regressor-based counterfactual fine-tuning of DSCMs, highlighting the improved effectiveness of our approach.
    \item We demonstrate the capability of generating realistic counterfactual chest radiographs. We also show promising early results on the application of modelling coronary artery disease progression.
\end{itemize}

\section{Method}
\subsection{Review of DSCM}
\paragraph{Structural Causal Models (SCMs)}\cite{pearl2009causality} are defined by a triplet $\langle U, A, F \rangle$, where $U = \{u_{i}\}_{i=1}^K$ represents a set of exogenous variables, $A=\{a_{i}\}_{i=1}^K$ a set of endogenous variables, and $F = \{f_{i}\}_{i=1}^K$ a set of functions satisfying $a_{k} \coloneqq f_{k}(\mathbf{pa}_{k}, u_{k})$, where $\mathbf{pa}_{k} \subseteq A \setminus a_{k}$ are the direct causes (or parents) of $a_k$. SCMs enable interventions through the do-operator, e.g., by modifying one or more parent variables. Counterfactual inference involves three steps: (i) Abduction: inferring exogenous noise from observed data; (ii) Action: applying an intervention, e.g. $do(a_{k} \coloneqq c)$; and (iii) Prediction: computing counterfactual outcomes using the modified model and the inferred posterior over the exogenous variables.

\paragraph{Deep Structural Causal Models (DSCMs)}  were introduced in \cite{pawlowski2020deep} and later refined in \cite{de2023high} for high-resolution counterfactual image generation. Given an image $\mathbf{x}$, let $\{a_{1}, \dots, a_{K-1} \} \supseteq \mathbf{pa}_{\mathbf{x}}$ be its \textit{ancestors}. Each low-dimensional attribute follows an invertible conditional normalizing flow, $a_{k} = f_{k}(u_{k}; \mathbf{pa}_{k})$, making abduction explicit and tractable. For high-dimensional variables like images, the generative mechanism is implemented via a Hierarchical Variational Autoencoder (HVAE). To generate a counterfactual image, we first infer the exogenous noise for the image, $\mathbf{z} \sim q_{\phi}(\mathbf{z} \mid \mathbf{x}, \mathbf{pa}_{\mathbf{x}})$, where $q_{\phi}$ is the HVAE encoder. Similarly, the exogenous noise for attributes is inferred as $u_{k} = f_k^{-1}(a_{k}; \mathbf{pa}_{k})$. We then perform an intervention by setting $a_{i} \coloneqq c$, allowing for modifications to multiple attributes simultaneously. Using the abducted noise $u_{k}$, we compute counterfactual parent values $\mathbf{\widetilde{pa}_{x}}$ and generate the counterfactual image, $\widetilde{\mathbf{x}} = g_{\theta}(\mathbf{z}, \mathbf{\widetilde{pa}_{x}})$.  


\subsection{Regressor-based Counterfactual Fine-tuning (Reg-CFT)}

Previous works~\cite{de2023high,xia2024mitigating,ibrahim2024semi} observed that likelihood-based HVAE training may cause \textit{ignored counterfactual conditioning}, where $\widetilde{\mathbf{x}}$ does not respect the intervened counterfactual parents $\mathbf{\widetilde{pa}_{x}}$. This can be mitigated with \textit{counterfactual fine-tuning (CFT)} \cite{de2023high,xia2024mitigating}. The key idea of CFT is to leverage pre-trained classifiers or regressors $q_{\xi}(\mathbf{{pa}_{x}} \mid {\mathbf{x}})$, and to optimise the HVAE parameters $\{\theta, \phi\}$ by maximising $\log q_{\xi}(\mathbf{\widetilde{pa}_{x}} \mid \widetilde{\mathbf{x}})$ while keeping $\xi$ frozen. This fine-tuning step encourages the DSCM to generate faithful counterfactual images that obey the intended interventions by enforcing $\mathbf{\widetilde{pa}_{x}}$ to be predictable from $\widetilde{\mathbf{x}}$. In this work, we refer to the CFT used in previous studies as Reg-CFT, as they employ pretrained regressors (or classifiers). For simplicity, we use the term \textit{regressor} to refer to both regressor and classifier throughout the paper.

While Reg-CFT has been demonstrated to be effective for \textit{subject-level interventions}, e.g., changing a subject's sex, it has not been tested on \textit{structure-specific interventions}, e.g., reducing the size of an organ in a medical image. We assess its effectiveness for structure-specific interventions by extracting \textit{areas} of structures as scalar parent variables for $\mathbf{x}$ using 2D medical images. As shown in~\cref{sec: experiments and results}, we find that Reg-CFT is not sufficient for these interventions, and produces undesirable global changes. A potential reason is that with Reg-CFT, there is insufficient guidance for DSCMs to capture the exact meaning of (scalar-valued) variables such as organ size, as the regressor could learn potential spurious correlations. 
For example, it is possible that DSCMs incorrectly learn that the variable \textit{left lung area} corresponds to the \textit{mean pixel intensities of left lung and heart} or some other spuriously correlated characteristics in the images. 
As such, it may be necessary to incorporate structural and spatially coherent information into CFT to better align the semantic meaning of scalar-valued, structure-specific variables.

\subsection{Segmentor-guided Counterfactual Fine-Tuning (Seg-CFT)}

\begin{figure}[b!]
    \centering
    \includegraphics[width=0.94\linewidth]{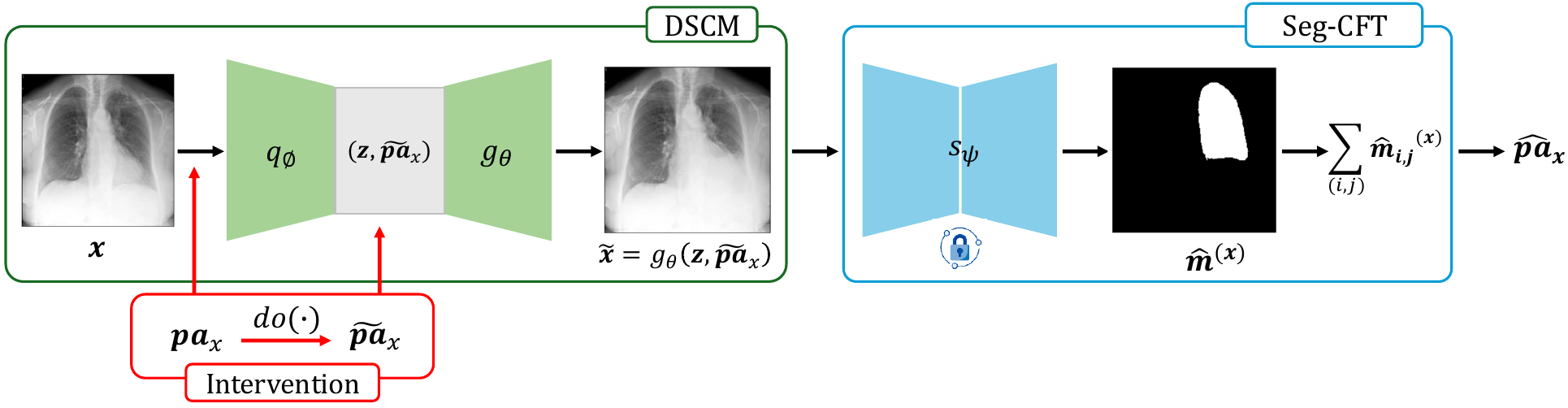}
    \caption{A schematic of the proposed Seg-CFT method, where we utilise pre-trained segmentors to guide counterfactual fine-tuning of DSCMs. }
    \label{fig: method}
\end{figure} 

For the rest of the paper, we focus on structure-specific interventions, i.e. changing areas of anatomical and disease-related structures in 2D images. To improve the effectiveness of this type of intervention, one potential approach is to rely on pixel-level segmentations to guide the generative models~\cite{perez2024radedit,alaya2024mededit}. This requires users to manually construct hypothetical, clinically plausible label maps, which is tedious and challenging. Additionally, incorporating label maps into a causal graph is inherently difficult, as the causal relationships between label maps and other variables may not be obvious. It is thus desirable to preserve the simplicity of intervening on scalar-valued causal variables while making the DSCM aware of the spatial context of these structure-specific variables. 

To this end, we introduce {Segmentor-guided Counterfactual Fine-Tuning (\textit{Seg-CFT})}, a novel approach that leverages a pre-trained segmentation model during counterfactual fine-tuning. For Seg-CFT, we retain all variables of interest as scalar-valued, similar to Reg-CFT, including both subject-level and structure-specific variables. This allows for tractable abduction, intervention and counterfactual reasoning for causal variables.  



When intervening on one or more variables with counterfactual parents \( \mathbf{\widetilde{pa}_x} \), we first predict the segmentation label maps $\mathbf{\widehat{m}}^{(\mathbf{x})}$ for different structures from  counterfactual images $\widetilde{\mathbf{x}}$ using segmentors $s_{\psi}$: 
$\mathbf{\widehat{m}}^{(\mathbf{x})} \sim s_{\psi}(\mathbf{\widehat{m}}^{(\mathbf{x})} \mid \widetilde{\mathbf{x}})$. 
Next, we compute the \textit{areas} of the structures by summing the pixel-level label probabilities of the predicted segmentations: $\mathbf{\widehat{pa}_{x}} = \sum_{(i,j)} \mathbf{\widehat{m}}^{(\mathbf{x})}_{i,j}$.


We then optimize the HVAE parameters \( \{\theta, \phi\} \) by minimizing the loss function \( l(\mathbf{\widehat{pa}_{x}}, \mathbf{\widetilde{pa}_{x}}) \), where \( l \) is designed for scalar-valued variables. A schematic of Seg-CFT is presented in \cref{fig: method}. The key difference between Seg-CFT and Reg-CFT is that Seg-CFT leverages pre-trained segmentors to \textit{indirectly} obtain \( \mathbf{\widehat{pa}_{x}} \), whereas Reg-CFT directly predicts \( \mathbf{\widehat{pa}_{x}} \) using regressors. With Seg-CFT, DSCMs must generate locally coherent and meaningful changes that affect the segmentation masks produced by the segmentor. This effectively forces the DSCM to learn the spatial context of the scalar-valued structure-specific variables. Notably, segmentors are used \textit{only} during training in Seg-CFT. Segmentors are not required during inference, and DSCMs can perform abduction, intervention, and counterfactual generation in the same manner as in previous works \cite{de2023high}.

\section{Experiments and results}
\label{sec: experiments and results}
We conduct experiments on two datasets: the publicly available PadChest~\cite{bustos2020padchest} and an internal coronary computed tomography angiography (CCTA) dataset~\cite{taylor2023}. Our evaluation compares the proposed Seg-CFT with Reg-CFT for structure-specific interventions, in particular by modifying the areas of targeted structures. For Reg-CFT, we pre-train ResNet-based regressors to predict the area of structures. For Seg-CFT, we pre-train U-Net segmentors using a Dice loss to produce 2D label maps. We evaluate counterfactuals via effectiveness~\cite{de2023high,monteiro2023measuring}, which assesses whether generated images obey the counterfactual parents, i.e. $d(\mathbf{\widehat{pa}_{x}}, \mathbf{\widetilde{pa}_{x}})$, where $d$ is a metric function. { The segmentor used for evaluation is not the same as the one used for fine-tuning; each is trained independently.}

\begin{table}[b!]
    \centering    \caption{\footnotesize{Quantitative evaluation of {effectiveness} of intervened and \unintervened{unintervened} variables. Best results are highlighted as \textbf{bold}. For PadChest, we measure MAPE ($\%$)}. For CCTA, we measure MAE (mm$^2$). Seg-CFT consistently outperforms other methods.}
    \footnotesize
    \begin{tabular}{@{}l|c|ccc|c|ccc}
        \hline \multirow{2}{*}{ \textsc{\footnotesize{CFT}}} &\multicolumn{4}{c|}{\footnotesize{\textsc{PadChest}}}         &\multicolumn{4}{c}{\footnotesize{\textsc{CCTA}}} 
        \\ 
        \cline{2-5}  
\cline{6-9}
         & \footnotesize{\textsc{Intervention}} & \footnotesize{\textsc{LLA}} & \footnotesize{\textsc{RLA}} & \footnotesize{\textsc{HA}} & \footnotesize{\textsc{Intervention}} & \footnotesize{\textsc{NCPA }} & \footnotesize{\textsc{CPA}} & \footnotesize{\textsc{LA}} \\
        \hline
        \footnotesize{No CFT} \cite{de2023high}  & \multirow{3}{*}{\footnotesize{do(LLA)}} & 16.1 & \unintervened{4.3} & \unintervened{6.7} & \multirow{3}{*}{\footnotesize{do(NCPA)}} & 17.27 & \unintervened{\textbf{3.45}} & \unintervened{5.75} \\
        \footnotesize{Reg-CFT}  \cite{de2023high} &  & 13.4 & \unintervened{2.2} & \unintervened{5.7} &  & 14.63 & \unintervened{6.83} & \unintervened{12.59} \\
        \footnotesize{Seg-CFT} (Ours) &  & \textbf{10.0} & \unintervened{\textbf{2.1}} & \unintervened{\textbf{5.2}} &  & \textbf{10.33} & \unintervened{3.98} & \unintervened{6.32} \\
        \hline
        \footnotesize{No CFT} \cite{de2023high}  & \multirow{3}{*}{\footnotesize{do(RLA)}} & \unintervened{3.7} & 16.9 & \unintervened{5.1} & \multirow{3}{*}{\footnotesize{do(CPA)}} & \unintervened{10.41} & 20.96 & \unintervened{7.97} \\
        \footnotesize{Reg-CFT} \cite{de2023high} &  & \textbf{\unintervened{1.3}} & 13.3 & \unintervened{4.6} &  & \unintervened{15.61} & 13.88 & \unintervened{12.10} \\
        \footnotesize{Seg-CFT} (Ours) &  & \unintervened{1.4} & \textbf{10.1} & \textbf{\unintervened{4.4}} &  & \textbf{\unintervened{10.40}} & \textbf{8.14} & \unintervened{7.49} \\
        \hline
        \footnotesize{No CFT} \cite{de2023high}  & \multirow{3}{*}{\footnotesize{do(HA)}}  & \unintervened{3.7} & \unintervened{4.0} & 14.2 & \multirow{3}{*}{\footnotesize{do(LA)}} & \unintervened{9.79} & \unintervened{\textbf{3.37}} & 14.84 \\
        \footnotesize{Reg-CFT} \cite{de2023high} &   & \textbf{\unintervened{1.3}} & \unintervened{2.1} & 11.8 &  & \unintervened{13.51} & \unintervened{4.98} & 12.66 \\
        \footnotesize{Seg-CFT} (Ours) &   & \textbf{\unintervened{1.3}} &\textbf{\unintervened{2.0}} & \textbf{8.5} &  & \unintervened{\textbf{9.50}} & \unintervened{3.51} & \textbf{9.68} \\
        \hline
    \end{tabular}
    \label{tab:final_combined_results}
\end{table}

\subsection{Study 1: Chest radiographs}

We begin by evaluating our method on chest radiographs from the PadChest dataset~\cite{bustos2020padchest}, intervening on the size of anatomical structures. We manually selected around 85k subjects, removing mislabeled or low-quality images. The resulting dataset consists of 61,714 images for training, 6,911 for validation and 17,123 for testing. All images were resized to $224 \times 224$ pixels. {We consider three structure-specific variables: left lung area (LLA), right lung area (RLA), and heart area (HA), with sex and age included as their parents.} The original PadChest data does not have segmentation masks. We use masks obtained with the pre-trained segmentation model of the \textit{torchxrayvison}~\cite{cohen2022torchxrayvision} package for left lung, right lung, and heart, respectively.

\begin{figure*}[b!]
    \centering
    \begin{subfigure}{1.0\textwidth}
    \centering
    \includegraphics[width=\textwidth]{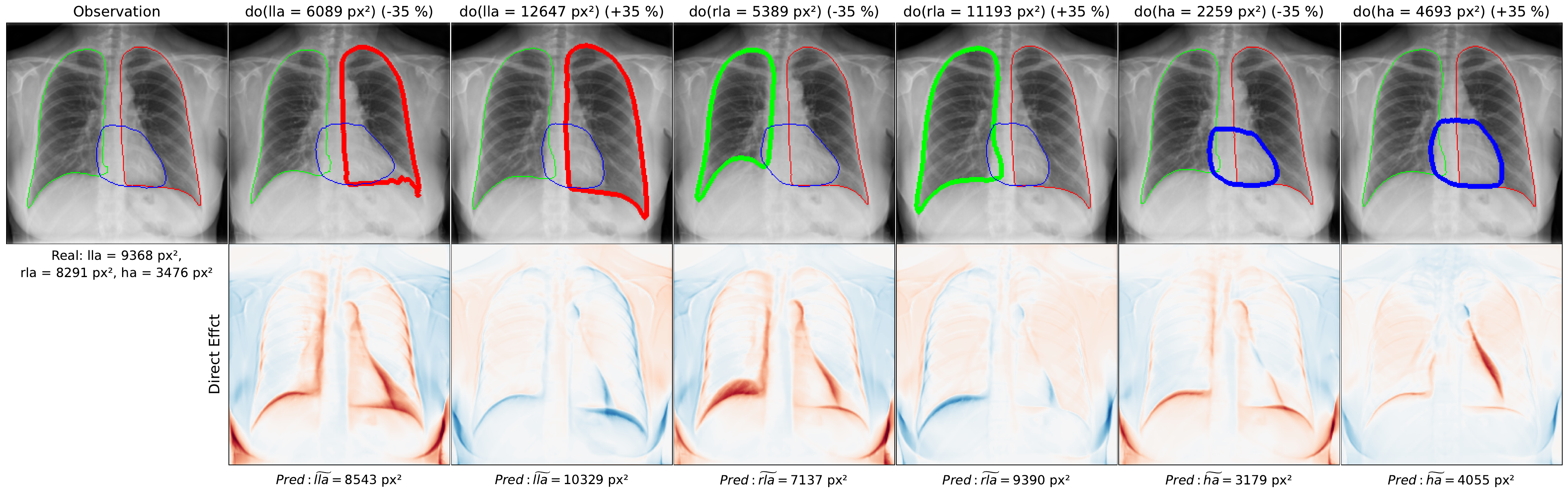}
    \caption{Reg-CFT}
    \label{fig:chest reg cft cf}
    \end{subfigure}
    \begin{subfigure}{1.0\textwidth}
    \centering
    \includegraphics[width=\textwidth]{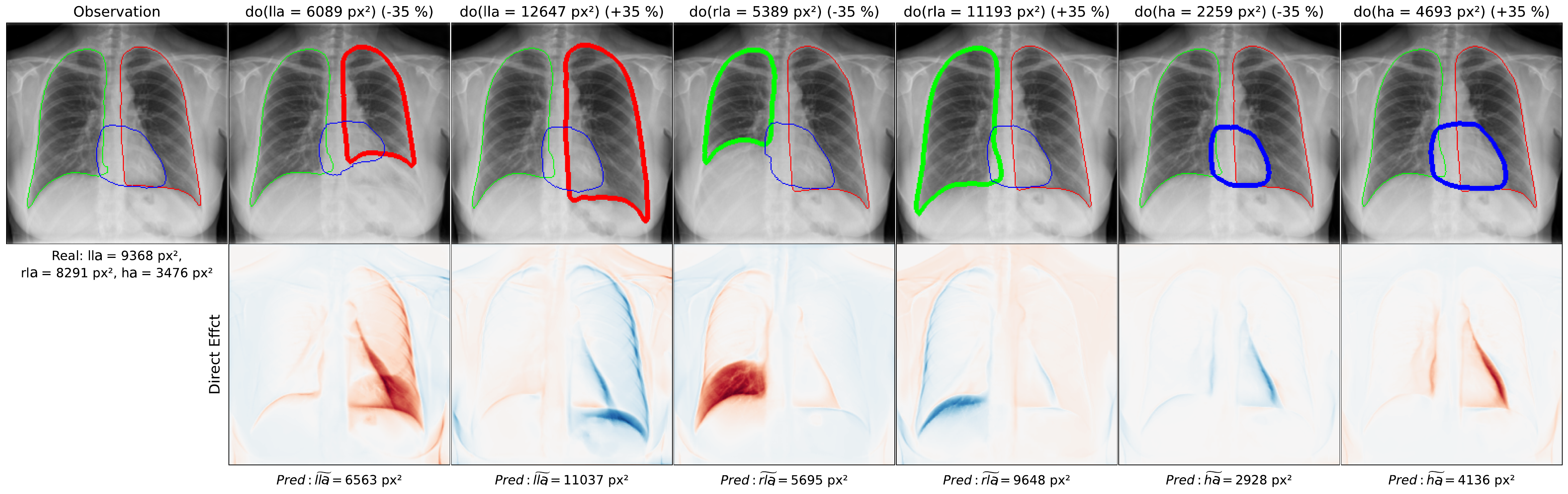}
    \caption{Seg-CFT}
    \label{fig: chest seg cft cf}
    \end{subfigure}
    \caption{Generated counterfactuals (CFs) with (a) Reg-CFT and (b) Seg-CFT. First rows show original image $\mathbf{x}$ and CFs $\mathbf{\widetilde{x}}$ with segmentations for {left lung} (red), {right lung} (green) and {heart} (blue). The intervened structure is highlighted with \textbf{thicker} lines. Second rows show direct effect of CFs, i.e. $\mathbf{\widetilde{x}}-\mathbf{x}$. We also report the predicted areas ($px^2$) by segmentors on the bottom. We observe that Seg-CFT produces more locally coherent and spatially consistent interventions.}
    \label{fig:chest CFs}
\end{figure*}

The quantitative effectiveness results are shown in \cref{tab:final_combined_results}, where we measure the mean absolute percentage error (MAPE) ($\%$) of LLA, RLA and HA for counterfactuals upon different interventions. We observe that without CFT, counterfactuals have the highest MAPE, highlighting the importance of CFT to mitigate ignored counterfactual conditioning. Seg-CFT achieves the best results with the lowest MAPE for all intervened variables.

\begin{figure*}[b!]
\centering
    \begin{subfigure}{0.95\textwidth}
    \centering
    \includegraphics[width=\textwidth]{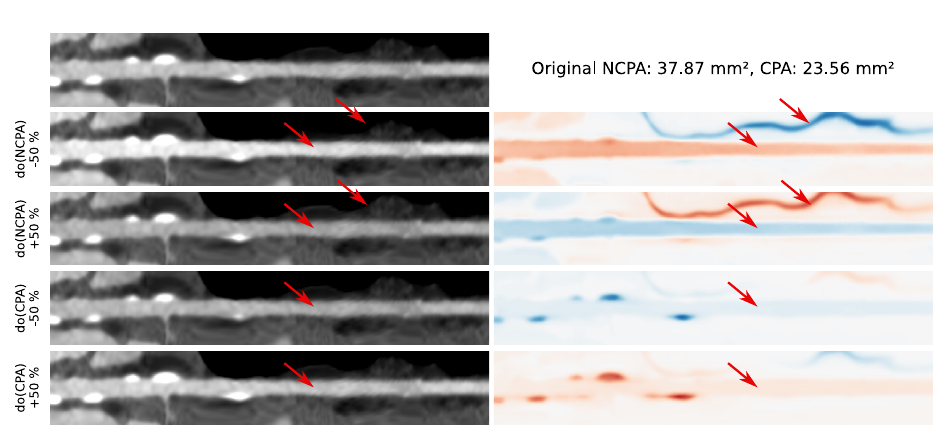}
    \caption{Reg-CFT}
    \label{fig: reg cft cf plaue}
    \end{subfigure}
    \begin{subfigure}{0.95\textwidth}
    \centering
    \includegraphics[width=\textwidth]{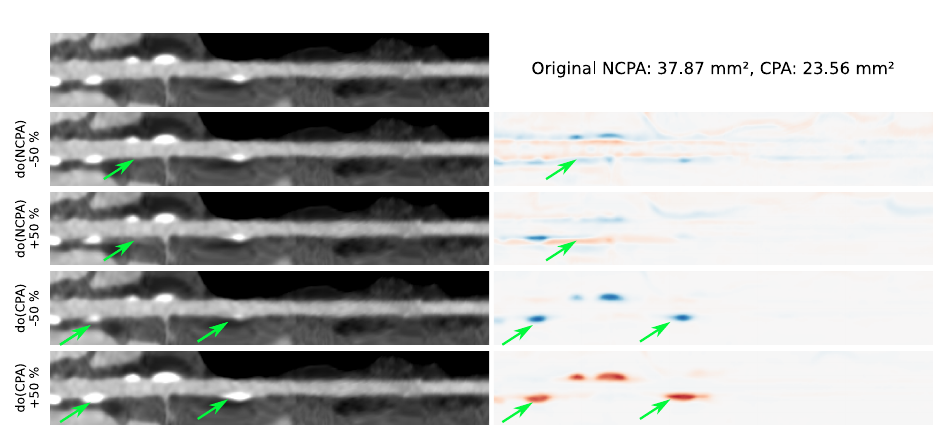}
    \caption{Seg-CFT}
    \label{fig:seg cft cf plaque}
    \end{subfigure}
    \caption{Generated CFs with (a) Reg-CFT and (b) Seg-CFT. Left columns show original image $\mathbf{x}$ and CFs $\mathbf{\widetilde{x}}$;  right columns show direct effect of CFs, i.e. $\mathbf{\widetilde{x}}-\mathbf{x}$. Seg-CFT produces more locally coherent and spatially consistent interventions. Green arrows indicate expected local changes in plaque with Seg-CFT, while red arrows highlight undesirable changes of non-target structures with Reg-CFT.}
    \label{fig:plaque CFs} 
\end{figure*}

Visual examples in \cref{fig:chest CFs} show that both {Reg-CFT} and {Seg-CFT} produce plausible counterfactuals, but we observe that {Reg-CFT} yields undesired effects outside the intervened structures. By contrast, with Seg-CFT, we observe more locally coherent and targeted changes, resulting in a more accurate intervention as reflected in the LLA, RLA and HA values predicted from counterfactuals. This suggests that with Seg-CFT, DSCMs obtain a better understanding of which part of an image should be changed upon structure-specific interventions.

\subsection{Study 2: Coronary artery disease}

CCTA is an important modality for the assessment of coronary artery disease (CAD), including evaluation of the composition and volume of atherosclerotic plaques. For the CCTA images, straightened curvilinear planar reformation (sCPR) was used to create 2D images from the longitudinal cross-section of the centerline of the left anterior descending (LAD) artery~\cite{kanitsar2002}. All images were sampled at a resolution of $0.25\!\times\! 0.25$ mm and cropped to $64\!\times\!384$ pixels.

Segmentation masks of the coronary lumen and plaque were also generated. To achieve this, 3D meshes of the coronary lumen and outer wall were sampled and rasterised in the 2D sCPR image plane, generating masks of the lumen, calcified plaque, and non-calcified plaque~\cite{narula2024,taylor2023patient}. A total of 18,433 CCTA images were used to generate samples, with 12,903 samples for training, 1,843 for validation, and 3,687 for testing. We consider three structure-specific variables: calcified plaque area (CPA), non-calcified plaque area (NCPA), and lumen area (LA). For simplicity, we assume that these are independent of each other.

The quantitative effectiveness of our approach is reported in \cref{tab:final_combined_results}, where we measure the mean absolute error (MAE) of NCPA, CPA, and LA in mm$^2$. Across all intervened variables, the proposed Seg-CFT achieves the best performance, followed by Reg-CFT. Notably, Reg-CFT results in significantly higher MAE for unintervened variables (indicated in \unintervened{gray} text). This is likely due to the regressors learning spurious correlations, which subsequently affect DSCMs during fine-tuning. The visual results in \cref{fig:plaque CFs} illustrate that Reg-CFT introduces unintended global effects across non-target structures. Note how interventions on NCPA affect the global intensity of the lumen area. In contrast, Seg-CFT yields much more localised effects, focusing specifically on the intervened structures. This demonstrates the advantage of incorporating segmentor-guidance in CFT.

\section{Conclusion}

By integrating pre-trained segmentation models during counterfactual fine-tuning, Seg-CFT enables locally coherent and targeted interventions while maintaining the simplicity of scalar-valued causal variables. Our experiments on PadChest for counterfactual chest radiographs, and on a CCTA dataset for simulating coronary artery disease progression, demonstrate that Seg-CFT outperforms regressor-based fine-tuning. Seg-CFT results in more targeted and structure-specific modifications while minimizing unintended global changes in unintervened regions. These findings highlight the importance of incorporating segmentation information to improve anatomical consistency. Future work should explore the causal relationship between structure-specific variables. We currently assume independence for simplicity. Practical applications of counterfactuals in areas such as disease progression modelling, treatment effect estimation, bias mitigation, and data augmentation should be explored. Extending Seg-CFT to 3D medical imaging, including volumetric CT and MRI scans, is another important direction that could unlock advanced counterfactual reasoning in high-dimensional data. Beyond controlling the area of anatomical structures, future research could investigate whether other characteristics, such as shape, location, or texture, can be explicitly modified within the counterfactual generation process, providing even greater flexibility in medical image synthesis.

\section*{Acknowledgments}
This research was funded by Heartflow. B.G. received support from the Royal Academy of Engineering as part of his Kheiron/RAEng Research Chair. B.G. and F.R. acknowledge the support of the UKRI AI programme, and the EPSRC, for CHAI - EPSRC Causality in Healthcare AI Hub (grant no. EP/Y028856/1). R.M. is funded through the European Union’s Horizon Europe research and innovation programme under grant agreement 10108030. R.R. is supported by the Engineering and Physical Sciences Research Council (EPSRC) through a Doctoral Training Partnerships PhD Scholarship.

\subsection*{Disclosure of interests}
M.S., A.S., E.P.A., S.G., K.P. and M.S. are employees of Heartflow. B.G. is part-time employee of DeepHealth. No other competing interests.

\bibliographystyle{splncs04}
\bibliography{Paper-1947}

\end{document}